\documentclass{article}
\usepackage[final]{staix_2026}

\usepackage[utf8]{inputenc}
\usepackage[T1]{fontenc}
\usepackage{hyperref}
\usepackage{url}
\usepackage{booktabs}
\usepackage{amsfonts}
\usepackage{amsmath,amssymb}
\usepackage{nicefrac}
\usepackage{microtype}
\usepackage{xcolor}
\usepackage{graphicx}
\usepackage{enumitem}

\newcommand{\suggest}[1]{\textcolor{black}{#1}}

\usepackage{tikz}
\usetikzlibrary{arrows.meta, positioning, shapes.geometric, fit, backgrounds, calc, decorations.pathreplacing}

\definecolor{inputblue}{HTML}{2B6CB0}
\definecolor{agentgreen}{HTML}{0077BB}  % colorblind-safe blue (Tol palette)
\definecolor{nudgeorange}{HTML}{C05621}
\definecolor{instrpurple}{HTML}{6B46C1}
\definecolor{strictred}{HTML}{C53030}
\definecolor{lenientgold}{HTML}{B7791F}
\definecolor{outgray}{HTML}{4A5568}
\definecolor{lambdabg}{HTML}{EBF5FB}  % light blue background for LAMBDA box
\definecolor{gradebg}{HTML}{FFFFF0}

\title{\textbf{Grading the Grader: Lessons from Evaluating an Agentic Data Analysis System}%
\thanks{Code and data available at GitHub:  
\url{https://github.com/TZstats-Columbia/STAI-X-Grade-The-Grader}.
}}

\author{%
  Tian Zheng \\
  Department of Statistics\\ Columbia University \\
  New York, NY \\
\texttt{tian.zheng@columbia.edu}
  \And
  Kai-Tai Hsu\\
  Department of Statistics\\
  Columbia University \\
  New York, NY \\
  \texttt{kh3400@columbia.edu}
}

\begin{document}

\maketitle

\begin{abstract}
Agentic data analysis systems produce rich outputs, including code, numerical results, and verbal diagnostics. This makes them more challenging to evaluate than single-turn LLM responses. It is therefore necessary to distinguish genuine disagreement between an agent's output and a ground-truth answer from grading artifacts. We investigate how reliably automated graders assess such a system and what strategies improve grading quality by applying LAMBDA, a multi-agent data-analysis system, on $153$ numerical QRData tasks from DSGym. We develop and evaluate a three-layer human-AI grading cascade: strict regex matching, LLM-based lenient grading, and snippet-based human inspection, which combines non-GenAI and GenAI strategies with different failure profiles. Both automated graders achieve $100\%$ \suggest{observed} precision \suggest{(0/70 false positives)}. The lenient grader's recall is $97\%$ against human labels. A keyword-anchored extraction pipeline raises the strict grader's recall by $60$ percentage points over a last-number heuristic; the lenient grader is architecturally parser-independent. An iterative nudge mechanism raises grading run success from $36\%$ to $97\%$ and lenient-pass rates from $16\%$ to $46\%$; comparing nudging with and without original-question re-injection shows that re-injection offers no benefit, confirming the nudge as an answer template cue. \suggest{We further observe in this case study that variable type is the task metadata field most consistently associated with grading pipeline dynamics and observed outcome grades.}
\end{abstract}

%==============================================================
\section{Introduction}
%==============================================================
\label{sec:intro}

Multi-agent LLM systems are increasingly deployed to automate data analysis workflows, and benchmarks such as DSGym \citep{nie2026dsgym} have made real progress toward systematic evaluation. One challenge is that, in addition to task completion, data analysis agents must also be evaluated on the correctness and rigor of their analytical results, while the \emph{graders} used to make this determination are themselves error-prone. An agentic system like LAMBDA \citep{sun2026lambda} generates rich, multi-step outputs, including code, execution logs, intermediate statistics, diagnostics, which are further interleaved with verbal comments. Extracting the intended answer from this output is non-trivial. In his discussion of LAMBDA, \citet{donoho2026discussion} calls for ``rigorous, reproducible benchmarks'' with scientific validity. More broadly, the veridical data science framework \citep{yukumbier2020, yuVeridicalDataScience2024} argues that any data-analytic pipeline must be evaluated along the axes of predictability, computability, and stability (PCS); applying these principles to agentic workflows \citep{rewolinski2025pcs} sharpens the question of what ``correct output'' means when the pipeline itself is stochastic and multi-step. 

In this paper, we study how reliably automated graders can assess agentic outputs and identify strategies to improve grading quality, focusing first on tasks with scalar ground truth. For example, DSGym’s default scorer, \texttt{exact\_match}, treats \texttt{"0.2"} and \texttt{"0.20"} as distinct, producing \emph{false negatives} and posing grading challenges \citep{nie2026dsgym}. This creates a structural difficulty for evaluating agentic AI on quantitative tasks: the richer the output, the harder it is to grade. We systematically address this \textit{auto-grading problem} for agentic data-analysis outcomes by examining agent outputs and their grades, quantifying how often graders fail even when agents agree with the ground truth, and developing practical strategies to improve grading quality and reliability. The {\bf contributions} of this paper are as follows. 
\begin{enumerate}[leftmargin=*, itemsep=1pt]
    \item \textbf{Evaluation framework with nudging.} A two-loop wrapper with per-turn instrumentation and iterative nudges adapts LAMBDA's conversational design to single-shot benchmarks, raising grading success from $36\%$ to $97\%$ and lenient-pass rates from $16\%$ to $46\%$.
    \item \textbf{Three-layer grading with human calibration.} A cascade of strict regex, LLM-based lenient grading, and snippet-based human checks combines non-GenAI and GenAI graders. Both automated graders reach $100\%$ observed precision; the lenient grader's recall is $97\%$ ($70/72$). Human audits of ~100-character snippets confirm 89\% of the auto-grading results and help spot cases that require further full review.
    \item \textbf{Parsing-grading interaction.} A keyword-anchored extraction pipeline improves strict-grader recall by $60$~pp over a last-number heuristic parser. The LLM-based lenient grader, run on full agent output, is parser-independent.
   \item \textbf{\suggest{Task covariates as interpretive signals.}} \suggest{To help interpret the observed grading dynamics, we analyze how they associate with simple task metadata (e.g., variable type) in this case study.} Variable type (categorical, continuous, or mixed) is the only task covariate that consistently aligns with grading dynamics and outcomes. \suggest{In this case study,} categorical tasks are harder to extract but exhibit a higher rate of agreement with the ground truth; continuous tasks are easier to grade but tend to produce more discrepancies between the agent’s outputs and the single ground truth.
   \item \textbf{Negative result: prompt re-injection.} Comparing nudge modes (answer-format only vs. plus original question) shows no benefit from re-injection on these short tasks: the nudge mainly cues answer format; the agent retains the answer internally.
\end{enumerate}

%==============================================================
\section{Background and Related Work}
%==============================================================
\label{sec:background}

\paragraph{LAMBDA.} LAMBDA \citep{sun2026lambda} is a multi-agent data-science assistant built around two core agents: the \emph{programmer}, which generates and executes code in response to user instructions, and the \emph{inspector}, which checks execution results for errors and issues revision suggestions when errors occur. This programmer-inspector loop iterates until the code runs without error or a maximum number of attempts is reached. Communication between agents is exclusively textual. Sun et al.\ benchmarked LAMBDA on classical tabular datasets for classification (accuracy) and regression (mean squared error), using five-fold cross-validation with results compared against manual R implementations; they also evaluated the system on high-dimensional, missing-data, image, and text datasets. We chose LAMBDA as our case study because it is open-source, light-weight, versatile, requires no fine-tuning, and has a key-value knowledge base that injects domain-specific code via in-context learning \citep{sun2026lambda}. It has demonstrated strong benchmark performance, making it a suitable testbed for studying grading methodology on a capable system.

\paragraph{DSGym and QRData.} DSGym \citep{nie2026dsgym} provides a unified evaluation harness over multiple data-science benchmarks. We use the QRData split, which contains $251$ problems; of these, $153$ have numerical (scalar) ground-truth answers. We carry out our investigation using these $153$ numerical tasks, each comprising a natural-language question, a tabular data file, and a single ground-truth scalar answer.

%==============================================================
\section{Methodology}
%==============================================================
\label{sec:methods}

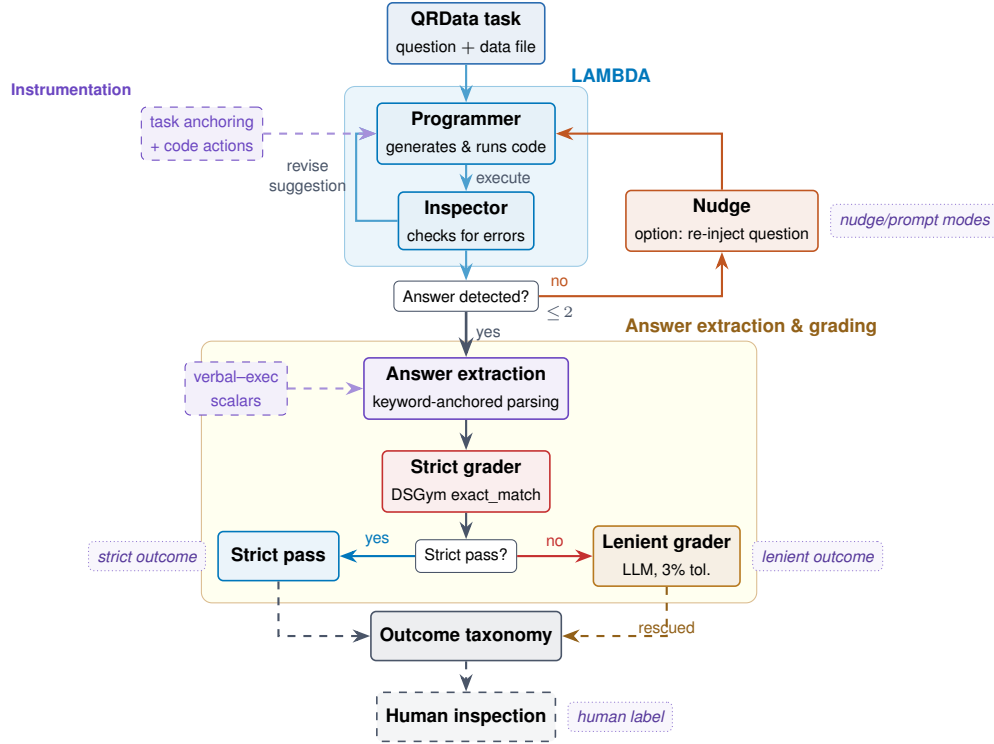
\begin{figure}[t]
\centering
\resizebox{0.95\columnwidth}{!}{% fig_eval_pipeline.tex — Evaluation methodology pipeline (TikZ)
% Usage: \input{fig_eval_pipeline.tex} inside a figure environment
\begin{tikzpicture}[
    >=Stealth,
    node distance=0.4cm and 0.7cm,
    every node/.style={font=\sffamily\scriptsize},
    box/.style={draw, rounded corners=2pt, minimum height=0.65cm,
                minimum width=1.6cm, align=center, line width=0.6pt},
    smallbox/.style={draw, rounded corners=2pt, minimum height=0.4cm,
                     minimum width=1.2cm, align=center, font=\sffamily\tiny,
                     line width=0.4pt},
    probe/.style={draw=instrpurple, fill=instrpurple!8, rounded corners=2pt,
                  minimum height=0.4cm, align=center,
                  font=\sffamily\tiny\color{instrpurple}, line width=0.4pt,
                  dashed},
    databox/.style={draw=instrpurple!70, fill=instrpurple!5, rounded corners=2pt,
                    minimum height=0.35cm, align=center,
                    font=\sffamily\tiny\itshape\color{instrpurple!80!black},
                    line width=0.4pt, densely dotted},
    scorebox/.style={fill=none, draw=none, align=center,
                     font=\sffamily\tiny\color{outgray}},
    arrow/.style={->, line width=0.8pt},
    thickarrow/.style={->, line width=1.0pt},
    dasharrow/.style={->, dashed, line width=0.8pt},
  ]

  %% === INPUT ===
  \node[box, fill=inputblue!10, draw=inputblue] (qrdata)
    {\textbf{QRData task}\\[2pt]{\tiny question $+$ data file}};

  %% === LAMBDA SYSTEM ===
  \node[box, fill=agentgreen!10, draw=agentgreen, below=0.5cm of qrdata] (prog)
    {\textbf{Programmer}\\[2pt]{\tiny generates \& runs code}};

  \node[box, fill=agentgreen!10, draw=agentgreen, below=0.35cm of prog] (insp)
    {\textbf{Inspector}\\[2pt]{\tiny checks for errors}};

  %% Arrows inside LAMBDA
  \draw[arrow, agentgreen!70] (qrdata) -- (prog);
  \draw[arrow, agentgreen!70] (prog) -- node[right, font=\sffamily\tiny, text=outgray]{execute} (insp);

  %% Inspector revision loop
  \draw[arrow, agentgreen!70] (insp.west) -- ++(-0.55,0) |- node[left, near start, font=\sffamily\tiny, text=outgray]{\shortstack{revise\\suggestion}} (prog.west);

  %% === NUDGE MECHANISM ===
  \node[box, fill=nudgeorange!10, draw=nudgeorange, right=1.2cm of insp] (nudge)
    {\textbf{Nudge}\\[2pt]{\tiny option: re-inject question}};

  %% Answer detection decision
  \node[smallbox, fill=white, draw=outgray, below=0.4cm of insp] (detect)
    {Answer detected?};

  \draw[arrow, agentgreen!70] (insp) -- (detect);

  %% No answer -> nudge
  \draw[arrow, nudgeorange] (detect.east) -- ++(0.55,0)
    node[above, midway, font=\sffamily\tiny, text=nudgeorange]{no}
    node[below, midway, font=\sffamily\tiny, text=outgray]{{\tiny$\leq\!2$}}
    -| (nudge.south);
  \draw[arrow, nudgeorange] (nudge.north) |- (prog.east);

  %% === DATA SIGNAL: nudge count ===
  \node[databox, right=0.15cm of nudge] (d_nudge) {nudge/prompt modes};

  %% === INSTRUMENTATION PROBE (left side, outside LAMBDA) ===
  \node[probe, left=1.5cm of prog] (p1) {task anchoring\\[2pt]+ code actions};

  \draw[dasharrow, instrpurple!60] (p1) -- (prog.west);

  %% Instrumentation label
  \node[above left=0.02cm and 0.0cm of p1, font=\sffamily\tiny\bfseries, text=instrpurple] {Instrumentation};

  %% === ANSWER EXTRACTION ===
  \node[box, fill=instrpurple!8, draw=instrpurple, below=0.6cm of detect] (extract)
    {\textbf{Answer extraction}\\[2pt]{\tiny keyword-anchored parsing}};

  \draw[thickarrow, outgray] (detect.south) -- node[right, font=\sffamily\tiny, text=outgray]{yes} (extract);

  %% Verbal--exec probe on extraction
  \node[probe, left=1.0cm of extract] (p2) {verbal--exec\\[2pt]scalars};
  \draw[dasharrow, instrpurple!60] (p2) -- (extract);

  %% === GRADING ===
  \node[box, fill=strictred!8, draw=strictred, below=0.4cm of extract] (strict)
    {\textbf{Strict grader}\\[2pt]{\tiny DSGym exact\_match}};

  \draw[arrow, outgray] (extract) -- (strict);

  %% Strict decision
  \node[smallbox, fill=white, draw=outgray, below=0.35cm of strict] (spass)
    {Strict pass?};

  \draw[arrow, outgray] (strict) -- (spass);

  %% Strict pass -> done
  \node[box, fill=agentgreen!8, draw=agentgreen, left=1.0cm of spass] (correct)
    {\textbf{Strict pass}};

  \draw[arrow, agentgreen] (spass.west) -- node[above, font=\sffamily\tiny, text=agentgreen]{yes} (correct);

  %% Strict fail -> lenient
  \node[box, fill=lenientgold!12, draw=lenientgold, right=1.0cm of spass] (lenient)
    {\textbf{Lenient grader}\\[2pt]{\tiny LLM, 3\% tol.}};

  \draw[arrow, strictred] (spass.east) -- node[above, font=\sffamily\tiny, text=strictred]{no} (lenient);

  %% === DATA SIGNALS: strict & lenient outcomes ===
  \node[databox, left=0.15cm of correct] (d_strict) {strict outcome};
  \node[databox, right=0.15cm of lenient] (d_lenient) {lenient outcome};

  %% === FINAL OUTPUT ===
  \node[box, fill=outgray!10, draw=outgray, below=0.5cm of spass] (taxonomy)
    {\textbf{Outcome taxonomy}};

  \draw[dasharrow, outgray] (correct.south) |- (taxonomy.west);
  \draw[dasharrow, lenientgold!80!black] (lenient.south) |- node[below, near start, font=\sffamily\tiny, text=lenientgold!80!black]{rescued} (taxonomy.east);

  %% === HUMAN INSPECTION ===
  \node[box, fill=outgray!6, draw=outgray, dashed, below=0.4cm of taxonomy] (human)
    {\textbf{Human inspection}};
  \draw[dasharrow, outgray] (taxonomy) -- (human);
  \node[databox, right=0.15cm of human] (d_human) {human label};

  %% === BACKGROUND BOXES ===
  \begin{scope}[on background layer]
    \node[fit=(prog)(insp), inner xsep=12pt, inner ysep=6pt, fill=lambdabg,
          draw=agentgreen!40, rounded corners=4pt, line width=0.4pt,
          label={[font=\sffamily\scriptsize\bfseries, text=agentgreen, anchor=south west, xshift=2pt, yshift=-2pt]above right:LAMBDA}] (lambdabox) {};

    \node[fit=(extract)(strict)(spass)(lenient)(correct), inner sep=6pt, fill=gradebg,
          draw=lenientgold!50, rounded corners=4pt, line width=0.4pt,
          label={[font=\sffamily\scriptsize\bfseries, text=lenientgold!80!black, anchor=south west, xshift=2pt, yshift=-2pt]above right:Answer extraction \& grading}] (gradebox) {};
  \end{scope}

\end{tikzpicture}}
\caption{Evaluation pipeline. Blue: LAMBDA's programmer--inspector loop. Purple (dashed): per-turn instrumentation probes. Orange: nudge mechanism (up to two retries if no scalar is detected). Yellow: dual grading---strict grader followed by LLM-based lenient grader with $3\%$ tolerance.}
\label{fig:pipeline}
\end{figure}

Figure~\ref{fig:pipeline} gives an overview of our evaluation pipeline. All experiments were conducted in a GitHub Codespace using 8 cores for task evaluation and 2 cores for result processing (no GPU); prior computational experiments utilized a MacBook Pro with M1 Max (no GPU). LAMBDA uses \texttt{gpt-5-mini} and \texttt{gpt-4.1-mini} via the OpenAI API; the lenient grader employed GPT-4o-mini (preliminary experiments with LLaMA-3 via Ollama produced similar results). \suggest{For transparency and reproducibility, we provide the wrapper-authored prompt templates used in Figure~\ref{fig:pipeline} (nudges and lenient grader) in Supplement~\ref{supp:prompt-transparency}.} The total pipeline time was ${\sim}120$ minutes per full run.

\label{sec:nudge-method}
\label{sec:per-step}

\noindent\textbf{Two-loop wrapper.}
LAMBDA was integrated into DSGym via a custom Python wrapper that imports LAMBDA in-process and calls its agents and workflow methods directly, with two nested loops. The \emph{inner loop} sends each QRData question to LAMBDA's programmer; if execution errors occur, LAMBDA's inspector issues revision suggestions and the programmer retries (up to $3$ iterations, matching Sun et al., Algorithm~1). The \emph{outer loop} handles missing or ambiguous answers: if the inner loop exits without error but no scalar is detected, or if the answer region contains more than three candidate numbers (Section~\ref{sec:answer-extraction}), the wrapper issues a \emph{nudge}, i.e., a follow-up prompt (see below for more details), and re-enters the inner loop (up to $2$ nudges). Each task has a $300$\,s timeout. Code and results JSON can be found in the project's GitHub repository.

\noindent\textbf{Nudge design.}
LAMBDA's programmer ends each turn with ``suggestions for next steps'' \citep[Figure~5]{sun2026lambda}. This offers natural interactions with users but may be interpreted as an incomplete answer in single-shot evaluation. The nudge bridges this mismatch. Every nudge includes an answer format (``state a single numerical answer; respond with ONLY the number''). The two modes differ only in whether the original question is re-injected: \textbf{Mode~A} (answer format $+$ original question) and \textbf{Mode~B} (answer format only). Comparing the two reveals whether recovery requires re-anchoring or merely an answer template cue.

\noindent\textbf{Per-step instrumentation.}
At every step the wrapper records instruction type, elapsed time, error status, extracted scalars, and the \emph{anchoring score} that evaluates keyword overlap between the agent turn and the original question: $\mathrm{Anchor}(t) = |K_q \cap \mathrm{tokens}(t)| / |K_q| \in [0,1]$, where $K_q$ is the set of content words from the question after stop-word removal (full field list in Supplement~\ref{supp:instrumentation}).

\noindent\textbf{Answer extraction.}
\label{sec:answer-extraction}
\label{sec:verbal-exec-method}
LAMBDA's response at each step is split into \emph{execution output} (text inside \verb|<pre>| blocks) and \emph{verbal text} (remainder after HTML removal). The extraction pipeline runs independently on both halves. The baseline \textit{heuristic parser} uses a very simple strategy: it scans the text and selects the final numeric value it encounters, treating that value as the predicted ``answer.'' The proposed improved \emph{keyword-anchored parsing pipeline} has three stages: (1) \emph{Answer-region identification:} the output is segmented into blocks, each scored by keyword overlap with the question, plus a bonus for answer-indicator phrases (``the answer is,'' ``therefore''); the highest-scoring block is selected. This block-level score is the \emph{answer-region confidence score}; if no block scores above $0.1$, the pipeline falls back to the last block, flagged as low-confidence. (2)~\emph{Multi-number extraction:} all numbers are extracted from the answer region, filtering out line numbers, array indices, and boolean-context values; if $1$--$3$ candidates remain, they pass to the grader. (3)~\emph{Ambiguity trigger}: if more than three candidates remain, an answer format nudge (as described above) is triggered.

As a consistency check, we compare the verbal and execution scalars via tolerance-based and LLM-based comparisons (Section~\ref{sec:verbal-exec}).

\noindent\textbf{Grading.}
\label{sec:grading}
Each task is graded at three levels. The \emph{strict} grader computes the distance from every candidate number (as described above) to the ground truth, selects the closest, and accepts it if $\mathrm{rtol}=0.02$ (with $\mathrm{atol}=10^{-6}$; $100\times$ for \% answers); technical details are in Supplement~\ref{supp:methods-ref}--\ref{supp:extraction-details}. Every strict failure is re-judged by a two-step \emph{lenient} process: GPT-4o-mini first extracts the intended answer semantically from the full output, then compares it to ground truth under $3\%$ tolerance. A task is lenient-pass if either grader accepts it. Finally, all $153$ tasks undergo \emph{human inspection}. \suggest{The human reviewer uses processed snippets (${\le}100$-character excerpts) as a triage aid, but the \textit{human label} is assigned at the end of the audit by reviewing the full JSON transcript for every task and applying the same $3\%$ tolerance rule used by the lenient grader when judging numerical agreement.} \suggest{Due to the scalar ground-truth setting of this case study, labeling ambiguity is minimal. Once a candidate answer is identified in the transcript, the match/non-match decision is usually straightforward under the specified tolerance rule. The primary difficulty lies in reliably locating the intended final answer within a noisy, multi-step output.} \suggest{Labeling was performed by a single annotator and was not blinded to the automated grades during snippet triage.} These human labels are used to evaluate the grading performance of the automated graders (detailed results in Section~\ref{sec:results}). \suggest{They are used only for post-hoc evaluation and exploratory error analysis of the grading pipeline (not for training, prompt tuning, or threshold selection in this study).}

\noindent\textbf{Outcome taxonomy and evaluation terminology.}
\label{sec:observation-vector}
We distinguish three levels of evaluation. \emph{Grading-run success} denotes whether the pipeline produces a scalar answer that can be graded. The \emph{outcome grade} is whether a particular grader (strict, lenient, or human) judges the extracted answer to agree with the ground truth. \emph{Ground-truth agreement} indicates whether the agent’s answer actually matches the ground truth, as verified by human inspection (i.e., the human label). We assess grader quality by comparing outcome grades to ground-truth agreement using precision and recall.

Each task is further classified by \emph{pipeline behavior}, independent of ground truth agreement: \texttt{clean} (scalar extracted without nudge or error recovery), \texttt{error\_recovered} (inspector resolved an execution error), \texttt{nudge\_recovered} (nudge required), or \texttt{failed} (no scalar after both loops). Ground truth agreement is assessed separately as described above. The per-task observation vector is detailed in Supplement~\ref{supp:instrumentation}.

\noindent\textbf{Task-characteristic analysis.}
\label{sec:clustering-method}
To test whether evaluation errors vary by task type, we examine four task-level covariates from the QRData metadata: variable type (categorical, continuous, or mixed), dataset dimension, whether statistical modeling is required, and analysis type (descriptive, inferential, or predictive/causal). Each covariate's association with grading pipeline dynamics and outcome grades is tested via a Chi-squared test of independence. \suggest{These associations are applied to help interpret the specific failure patterns observed in this case study.} (Section~\ref{sec:clustering}).

%==============================================================
\section{Results}
%==============================================================
\label{sec:results}

\noindent\textbf{Ground-truth agreement.}
Human inspection established grading ground truth for all $153$ tasks: $72$ matched the ground truth ($47.1\%$), $81$ did not. \suggest{The human label was assigned after full-transcript review for every task.} \suggest{We also report the performance of a snippet-only triage as a practical audit aid: using only the processed snippets for the tasks, the snippet-stage judgment agreed with the final label in $136/153$ cases ($89\%$). In our error analysis, snippet review was more reliable for confirming non-matches than for confirming matches. Two common sources of snippet-stage disagreement were (i) truncated outputs that captured LAMBDA UI placeholders and (ii) verbose reasoning text being included in the excerpt, which can obscure where the final answer appears. For these reasons, cases flagged as potential false positives by snippet review should be followed up with full-transcript inspection.}

\noindent\textbf{Grader performance.}
\textit{Outcome grades} produced by automated graders reflect both the agent's ground-truth agreement and the grader's ability to detect it. Table~\ref{tab:grader-perf} evaluates each grader configuration against the human labels. All configurations achieve $100\%$ observed precision (no false positives), \suggest{with a 96\% lower bound based on the \textit{rule of three} \citep{tuyl2009rule}}. Recall varies widely: the strict grader with a heuristic parser recovers only $26\%$ of ground-truth matches, while the keyword-anchored parser raises this to $86\%$ and the lenient grader reaches $97\%$. Without nudging, both graders suffer large recall losses because the agent's first-pass output is too verbose for reliable extraction.

\begin{table}[t]
\centering
\caption{Grader performance against human labels (Mode~A, $n=153$; human labels: $72$ match, $81$ non-match). TP = true positives; FP = false positives; FN = false negatives; TN = true negatives.}
\label{tab:grader-perf}
\label{tab:accuracy}
\begin{tabular}{lcccccc}
\toprule
\textbf{Grader configuration} & \textbf{Grading succ.} & \textbf{TP} & \textbf{FP} & \textbf{FN} & \textbf{TN} & \textbf{Recall \suggest{(s.e.)}} \\
\midrule
\multicolumn{7}{l}{\emph{With nudge}} \\
\quad Strict, heuristic parser          & 149 & 19 & 0 & 53 & 81 & 26\% \suggest{(5\%)} \\
\quad Strict, KW-anchored parser        & 149 & 62 & 0 & 10 & 81 & 86\% \suggest{(4\%)}\\
\quad Lenient grader                     & 149 & 70 & 0 & 2  & 81 & 97\% \suggest{(2\%)}\\
\addlinespace
\multicolumn{7}{l}{\emph{Without nudge (reconstructed from pre-nudge output)}} \\
\quad Strict, KW-anchored parser        & 55  & 17 & 0 & 55 & 81 & 24\% \suggest{(5\%)}\\
\quad Lenient grader                     & 55  & 25 & 0 & 47 & 81 & 35\% \suggest{(6\%)}\\
\bottomrule
\end{tabular}
\end{table}

\begin{figure}[h]
\centering
\includegraphics[width=\textwidth]{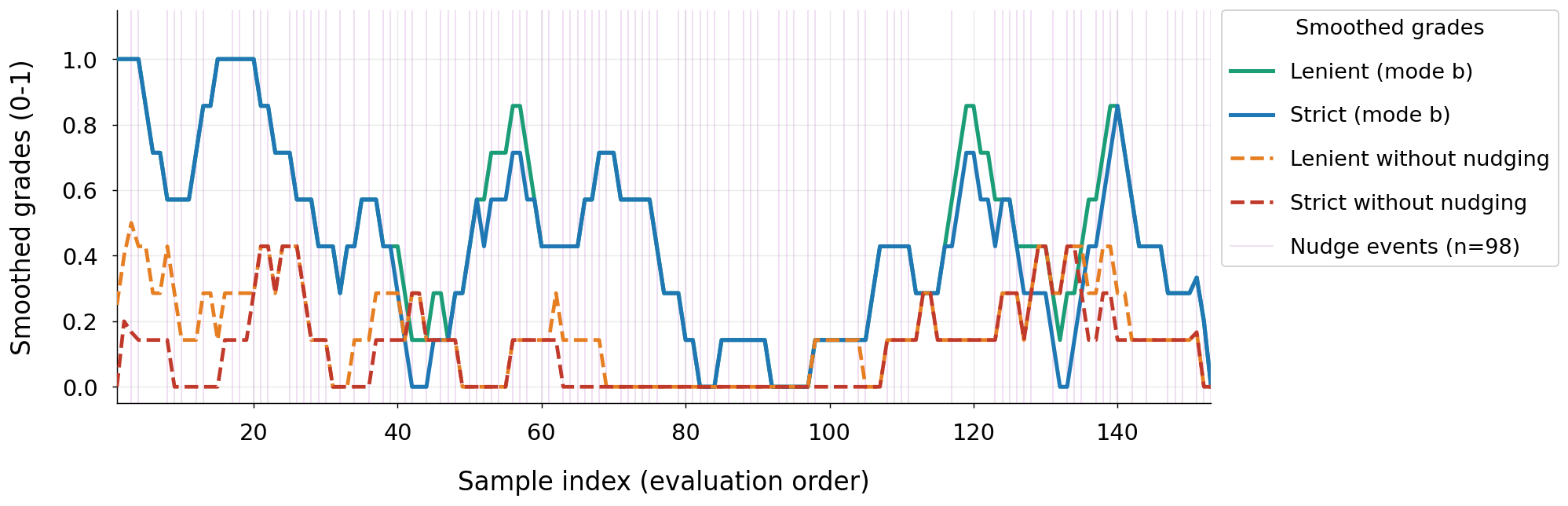}
\caption{Smoothed outcome grades across all $153$ tasks (Mode~B, in evaluation order). Vertical grey lines mark nudge events.}
\label{fig:grading-panel}
\end{figure}

Figure~\ref{fig:grading-panel} visualizes these differences across the $153$ tasks in evaluation order: the strict grader without nudging (red dashed line) rarely detects matches; The LLM-lenient grader (orange dashed line) without nudging can detect more matched cases but is limited by LAMBDA's innate outputs (see below for more discussion). Adding nudging and keyword-anchored parsing (solid blue) substantially increases the detection rate; and the lenient grader (green) recovers most of the remaining true positives. The gap between the blue and green curves reflects cases where the agent's answer matches the ground truth but is formatted in a way that strict matching cannot capture.

\smallskip
\noindent\textbf{Nudging improves both grading run success and outcome grades.}
$98$ tasks ($64\%$) triggered at least one nudge (Figure~\ref{fig:grading-panel}). Among these, $94$ recovered a more gradable answer. This high nudge effectiveness rate reflects a design mismatch between LAMBDA’s conversational interface and DSGym QRData’s evaluation framework's single-shot format. The nudge provides the answer template cue that the benchmark format lacks. Case study~2 (Supplement~\ref{supp:case-nudge}) illustrates the most common pattern: the agent computes $\lambda = 4.4$ correctly on the first substantive turn but appends numbered suggestions that inject answer-irrelevant numbers and trigger the ambiguity condition. Two nudges elicit a clean scalar.

To systematically quantify this, we reconstruct a no-nudge baseline from pre-nudge step histories (Table~\ref{tab:grader-perf}, lower panel). Without nudging, grading run success drops to $55/153$ ($36\%$) and the lenient grader’s recall falls from $97\%$ to $35\%$. Of the $94$ nudge-recovered tasks, the lenient grader returns \texttt{unmatched} on $78$ ($83\%$) of their pre-nudge outputs: the first-pass response is too verbose for even an LLM grader to locate a scalar answer.

\suggest{Table~\ref{tab:prenudge-presence} further separates cases where a candidate scalar answer was already detectable pre-nudge from cases where it was hard to find. Only $16/94$ ($17\%$) contain a detectable scalar answer in the raw pre-nudge output; among these, $8$ are already correct pre-nudge and $8$ contain a scalar that is not correct under the $3\%$ tolerance rule. Thus, nudging most often acts as an answer-format cue that makes the final scalar explicit and gradable, while only occasionally prompting recomputation that changes an already-present (but incorrect) scalar.}

\begin{table}[t]
\centering
%\color{blue}
\caption{Evaluation of the pre-nudging transcripts by the lenient grader on nudge-recovered tasks (Mode~A; $n=94$). ``Answer present” indicates that a candidate scalar answer was detected in the pre-nudge transcript but was not extracted by the heuristics-based parser; “answer absent” indicates that no such scalar answer was detected in the pre-nudge transcript. Answer correctness is evaluated under the $3\%$ tolerance rule.
}
\label{tab:prenudge-presence}
\begin{tabular}{lrr}
\toprule
Category & $n$ & \% of nudge-recovered \\
\midrule
Answer not detected (lenient scalar not found pre-nudge) & 78 & 83.0\% \\
Answer detected in raw output (total)               & 16 & 17.0\% \\
\quad of which detected but not match GT    & \textit{8} & \\
\quad of which match GT pre-nudge           & \textit{8} & \\
\midrule
Total nudge-recovered & 94 & 100\% \\
\bottomrule
\end{tabular}
\end{table}

\smallskip
\noindent\textbf{Keyword-anchored parsing greatly improves strict grading.}
\label{sec:parser-comparison}
The keyword-anchored pipeline raises the strict grader's recall from $26\%$ to $86\%$ against human labels (Table~\ref{tab:grader-perf}). This reveals a parsing-grading asymmetry: strict grading depends entirely on the parsing pipeline, while the LLM-lenient grader performs its own semantic extraction from the full agent output and is architecturally independent of parsing. The strict-lenient gap \emph{widens} from $0.7$~pp (heuristic) to $5.2$~pp (keyword-anchored): the better parser captures easy cases via strict matching, leaving the lenient grader with only the genuinely hard extraction problems. 

Case study~1 (Supplement~\ref{supp:case-clean}) illustrates the fragility of the strict grader: a \texttt{data.head()} call floods the output with numbers. The strict grader failed to find a match under Mode A but succeeded under Mode B, while the lenient grader succeeded under both modes. 

The answer-region confidence score further provides an internal quality indicator: tasks in the highest-confidence band ($\ge 0.75$) achieve a $53\%$ average strict grade, roughly double the rate at lower confidence levels (Figure~\ref{fig:region-confidence} in Supplement~\ref{supp:extraction-details}).

This parsing-grading asymmetry does not diminish the practical value of keyword-anchored \textit{strict grading}. LLM-based lenient grading inherits well-documented LLM limitations such as hallucination, context-length sensitivity, and output drift \citep{ji2026overview, gu2025attention}, any of which could generate false positives difficult to detect at scale. The keyword-anchored pipeline, by contrast, is deterministic and non-generative. In other words, it can only extract numbers that actually appear in the output. Our evaluation thus deliberately combines \emph{both} strategies: the non-GenAI tier (nudging plus keyword-anchored parsing) provides a first-pass grader with no hallucination risk; the GenAI tier (lenient grading on full output) recovers cases that deterministic extraction cannot handle. Both tiers are valuable precisely because they fail in different ways.

\smallskip
\noindent\textbf{\suggest{Variable type is the most consistently associated task metadata field in this case study.}}
\label{sec:clustering}
Of four task-level covariates tested (variable type, dataset dimension, modeling requirement, analysis type), only \emph{variable type} \suggest{is observed to have} a consistent association across both modes and both targets (Table~\ref{tab:variable-type}), \suggest{in this case study}. 

\begin{table}[t]
\centering
\caption{Association between variable type and outcome grades ($n=153$ per mode). Cram\'{e}r's~$V$ with chi-squared $p$-values.}
\label{tab:variable-type}
\begin{tabular}{lcccc}
\toprule
\textbf{Target} & \textbf{Mode~A $p$} & \textbf{Mode~A $V$} & \textbf{Mode~B $p$} & \textbf{Mode~B $V$} \\
\midrule
Strict binary  & $0.003$ & $0.28$ & $0.015$ & $0.23$ \\
Lenient binary & $<0.001$ & $0.33$ & $0.011$ & $0.24$ \\
\bottomrule
\end{tabular}
\end{table}

\begin{figure}[t]
\centering
\includegraphics[width=\textwidth]{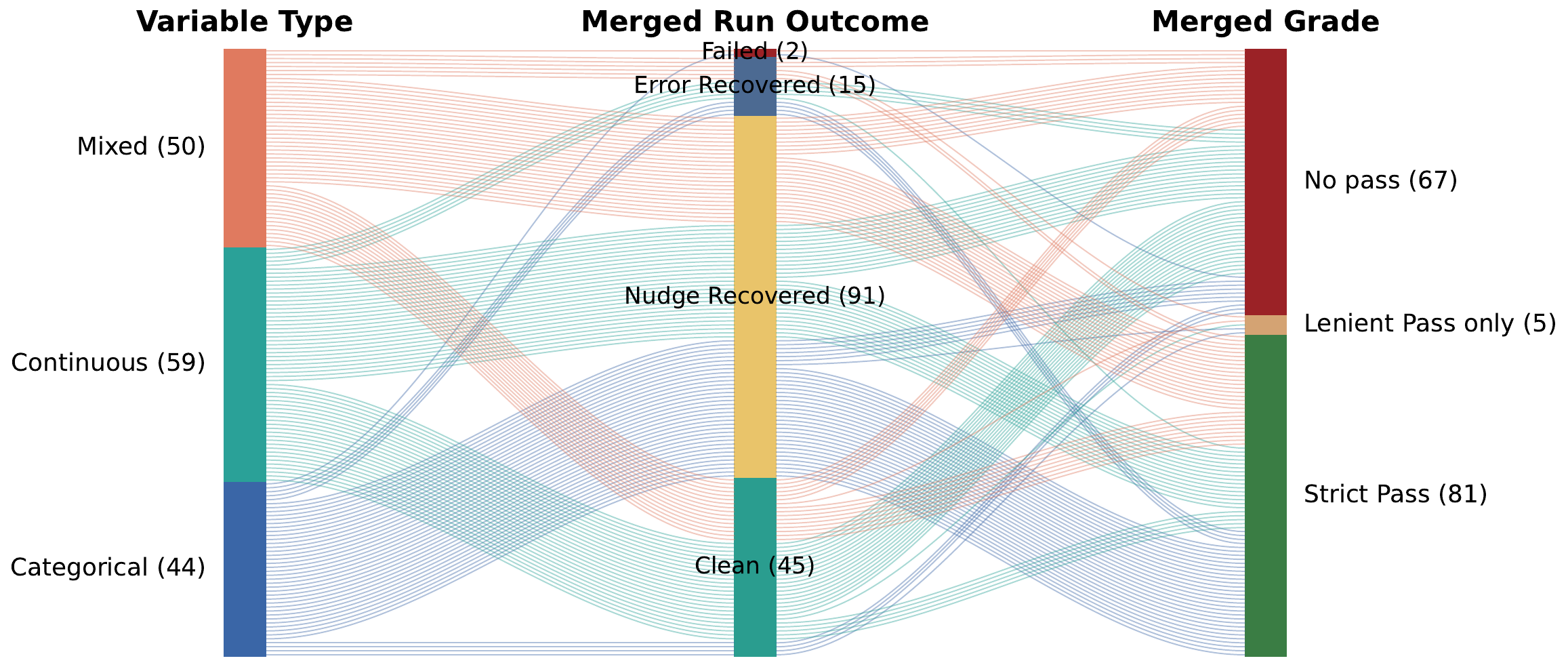}
\caption{Sankey diagram of variable type $\to$ pipeline outcome $\to$ grading result (merged across Mode~A and Mode~B). Categorical tasks flow predominantly through nudge recovery to pass grades; continuous tasks spread more evenly across outcomes but concentrate among non-match grades.}
\label{fig:sankey}
\end{figure}

Figure~\ref{fig:sankey} shows how the pipeline dynamics differ by variable type. \emph{Categorical} tasks ($n=44$) are the most nudge-dependent ($35/44=79\%$ require nudging) yet achieve the highest strict-pass rate ($80\%$): the agent's reasoning is sound, but categorical outputs, such as frequency tables, flood the initial response with numbers that challenge deterministic extraction. The same reason also explains why the lenient grade shows a stronger association with variable type than the strict grade (Table~\ref{tab:variable-type}). LLM grading and nudging both help overcome the verbose output style that categorical tasks produce. \emph{Continuous} tasks ($n=59$) present the opposite profile: the highest clean rate ($25/59 = 42\%$) but the lowest strict-pass rate ($ 76\%$), producing \emph{``false'' clean} runs where the agent outputs a plausible but unmatched number without triggering the nudge. Many continuous tasks require more data processing and analysis decisions, where subtle differences in methodological choices yield different scalars. Case study~3 (Supplement~\ref{supp:case-failed}) illustrates a limitation of the match/non-match dichotomy: the agent computes a one-sided $p$-value ($0.006$) instead of the two-sided value ($0.012$), a methodological decision discrepancy that no scalar comparison can diagnose and that a richer grading taxonomy (e.g., partial credit for correct methodology with a different specification) could address. This contrast highlights that \emph{tasks that are hard to grade are not the same as tasks that are hard to execute}. An evaluation framework for data analysis agentic systems must address both axes independently.

\smallskip
\noindent\textbf{Prompt re-injection does not help.}
\label{sec:mode-comparison}
The two nudge modes produce very similar outcome grades (Mode~A: $40.5\%$ strict, $45.8\%$ lenient; Mode~B: $43.1\%$, $46.4\%$), confirming that re-injecting the question offers no benefit on these short-run tasks. Re-injection may matter more on longer tasks where genuine drift accumulates. Full comparison tables are in Supplement~\ref{tab:mode-comparison-supp}.

The prompt structure of QRData tasks may partially explain this pattern. Each prompt contains an \emph{analysis instruction} followed by a \emph{reporting instruction} (e.g., ``Please round to the nearest hundredth''). The agent stays stateful with respect to the analysis but loses track of the reporting format. Nudging succeeds because it reminds the agent of the \emph{reporting} step. Re-injecting the full task prompt adds no value because the agent has not forgotten what to compute, only what to report.

\begin{figure}[h]
\centering
\begin{minipage}[t]{0.50\textwidth}
\centering
\includegraphics[width=\textwidth]{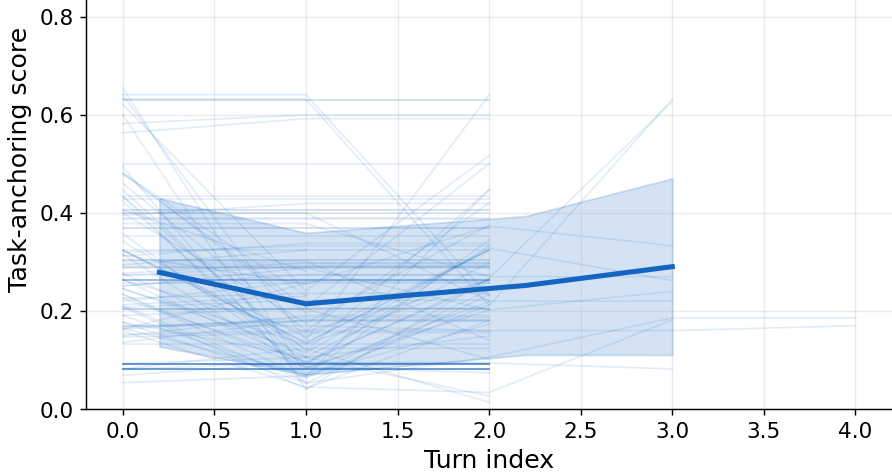}
\centerline{\small (a) Mode~A: answer format + question re-injection}
\end{minipage}\hfill
\begin{minipage}[t]{0.50\textwidth}
\centering
\includegraphics[width=\textwidth]{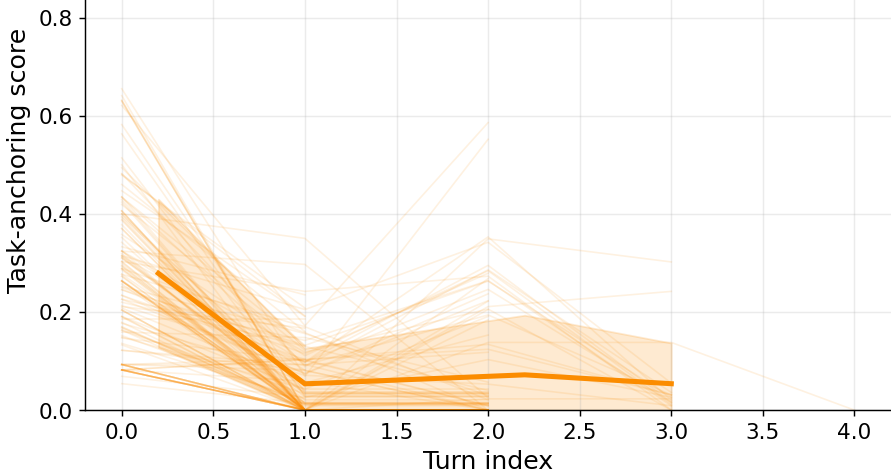}
\centerline{\small (b) Mode~B: answer format only}
\end{minipage}
\caption{Task-anchoring trajectory by turn ($n=153$ per mode). Faint lines: individual tasks, colored by outcome category. Bold line: mean anchoring score. Mode~A (left) maintains anchoring via question re-injection. Mode~B (right) decays monotonically, yet achieves comparable outcome grades ($43.1\%$ vs.\ $40.5\%$ strict). \medskip}
\label{fig:trajectory}
\label{fig:main_traj}
\end{figure}

The per-turn anchoring trace (Figure~\ref{fig:trajectory}) reveals the mechanistic difference. Mode~A maintains anchoring around $0.2$--$0.3$ via question re-injection; Mode~B decays monotonically toward zero. Yet ground-truth agreement is nearly identical, showing that low anchoring does not indicate task deviation. The agent's code uses different terminology from the prompt (variable names and function calls vs.\ the natural-language question). Genuine drift does occur (e.g., the one-sided $p$-value error in Case~3), but vocabulary shift and genuine drift are confounded. A keyword-based score cannot distinguish correct domain-specific reasoning from quiet deviation. Disentangling these two phenomena is a central difficulty in grading agentic data-analysis workflows.

\noindent\textbf{Verbal-execution divergence.}
\label{sec:verbal-exec}
Of the 153 tasks, the heuristic parser flagged 46 verbal–execution discrepancies, whereas the keyword-anchored parser reduced this to 0. The discrepancies identified by the heuristic parser expose various parsing edge cases; further details are provided in Supplement~\ref{supp:disagreements}.

\smallskip
\noindent\textbf{Stability.}
A small stability experiment ($40$ runs on a single task under four conditions: replication and prompt perturbation $\times$ Mode~A and Mode~B) yielded $38/40$ ($95\%$) strict-pass outcomes. The two failures occurred in Mode~B replication runs, where the extractor locked onto a spurious number in a \texttt{clean} run (no nudge was triggered). Ground-truth agreement on this benchmark appears highly stable across runs, though broader stability characterization is left for future work.

%==============================================================
\section{Conclusion and Discussion}
%==============================================================
\label{sec:conclusion}
\label{sec:discussion}

\paragraph{Summary of findings.}
This paper investigated the reliability of automated grading for agentic data analysis outputs, using LAMBDA on $153$ numerical QRData tasks as a case study. Outcome grades are a composite of the agent’s ground-truth agreement and the grader’s ability to detect it. Disentangling the two requires human calibration. Five findings stand out: (1)~nudging raises grading run success from $36\%$ to $97\%$ and grading recall from $35\%$ to $97\%$; (2)~a three-layer grading cascade combining non-GenAI and GenAI strategies achieves $100\%$ observed precision and up to $97\%$ recall; (3)~keyword-anchored parsing raises the strict grader’s recall from 26\% to 86\%, while the lenient grader is architecturally parser-independent; (4)~variable type is \suggest{observed as} the only covariate consistently associated with both grading pipeline dynamics and outcome grades, with categorical and continuous tasks exhibiting different profiles \suggest{ in this case study}; (5)~re-injecting the original question during nudging offers no benefit, suggesting that agents remain stateful for short tasks between iterations.

\smallskip
{\bf Need for effective human inspection.}
Because outcome grades conflate agent errors with grading errors, developers who rely on uncalibrated automated metrics may end up optimizing the wrong factors. We propose calibrating each automated metric against human labels on a representative subsample and reporting this audit alongside the metric. A snippet-based fast-review protocol achieves 89\% agreement with the final human label, providing a cost-effective method for human inspection.

\smallskip
{\bf Beyond single-scalar ground truth.}
This paper focuses on tasks with a single scalar ground truth, which is a setting that is narrow but rigorous, scaling to hundreds of tasks without subjective rubrics. Yet even in this clean setting, a scalar disagreement does not necessarily mean the agent produced a wrong analysis; it may reflect a defensible alternative specification (e.g., different handling of missing values or different method choices, such as one-sided vs. two-sided tests, as in Case Study~3). This ambiguity is especially pronounced for continuous-variable tasks, which carry the highest rate of ground-truth disagreement in our data.

In practice, evaluating data analysis must go further. As \citet{donoho2026discussion} emphasizes, assessing data analysis agents ultimately requires scrutinizing their intermediate reasoning steps—data cleaning, transformation logic, statistical specification, and interpretation. These steps make such agents particularly difficult to evaluate: small methodological differences can yield plausible yet materially different results, the ground truth may legitimately admit multiple correct answers, and comparing outputs across multiple agentic runs further compounds the challenge. Yet these scenarios share a common core difficulty: reliably processing, extracting, and comparing agent outputs.

The grading strategies developed in this study, such as nudging agents to produce clean outputs, using keyword-anchored parsing to locate answers, and applying LLM-based lenient grading to accommodate formatting variation, directly target this difficulty and extend readily to richer evaluation settings. They may be added to a human-AI collaboration workflow for evaluating agentic AI data analysis systems by automating grading layers, triaging and classifying at scale, and enabling human inspection that calibrates judgments and catches what automation misses.

\smallskip
{\bf Limitations.}
Our study focuses on the full set of $153$ numerical QRData tasks and \suggest{a single agentic system} (LAMBDA). This breadth supports exploratory, task-level analysis of grading behavior but limits our ability to make broad claims about data analysis in general. \suggest{Accordingly, our contribution should be read as evidence from this case study that the evaluation must disentangle grader artifacts from true agent disagreement, rather than as a general claim about task difficulty for agentic AI systems across data-analysis settings.} Our keyword-anchored extraction pipeline depends on lexical overlap between the question and answer region, and may underperform when correct answers are expressed using substantially different terminology or notation. A small stability experiment ($40$ runs) indicates that both the agent and grading cascade are highly reproducible under our settings, yet a more systematic robustness study (e.g., across prompts, models, and random seeds) remains future work. The observed $1.3\%$ false-negative rate is specific to the QRData benchmark and our configuration, and may not transfer to tasks with more complex answer formats or weaker ground truth. \suggest{Because the agent and lenient grader are drawn from the same model family, systematic shared-family biases are possible; cross-family grading is an important direction for further exploration.} All code, grading artifacts, and task configurations are released at the project repository; an online supplement (S.1--S.9) documents implementation details, additional analyzes, and ablations.

\smallskip
Evaluating agentic data analysis systems is itself a data-analytic task that requires an exploratory, iterative, and multi-perspective statistical workflow. By constructing a human-AI grading cascade, calibrating automated metrics against human labels, and analyzing how task characteristics influence both grading pipeline dynamics and final grades, this work provides reusable evaluation strategies along with design insights for practitioners building or benchmarking data analysis agents. More broadly, our framework illustrates how principled measurement, careful uncertainty quantification, and systematic ablation can turn the evaluation of complex AI systems into a transparent and reproducible empirical exercise.
%==============================================================
\bibliographystyle{plainnat}
\bibliography{staix-evalAI}

\newpage
% supplement-body.tex  —  included by main.tex via \input{supplement-body}
% Also used by supplement.tex (standalone compilation) via \input{supplement-body}
%
% Do NOT add \documentclass, \begin{document}, or \end{document} here.

% Restart page number and line numbers, with S‑prefix for pages
\setcounter{page}{1}% restart page numbering
\renewcommand{\thepage}{S\arabic{page}}% page numbers: S1, S2, ...

% If you use the "lineno" package and want line numbers restarted
% (and optionally prefixed with S as well):
%\usepackage{lineno}
%\linenumbers
%\setcounter{linenumber}{1}% restart line numbers
% Optional: prefix line numbers with S (comment out if not desired)
%\renewcommand\thelinenumber{S\arabic{linenumber}}

% Your existing supplement counters
\setcounter{section}{0}
\renewcommand{\thesection}{S.\arabic{section}}
\renewcommand{\thetable}{S.\arabic{table}}
\renewcommand{\thefigure}{S.\arabic{figure}}
\setcounter{table}{0}
\setcounter{figure}{0}

\begin{center}
{\Large {\bfseries Online Supplement} \\ \vspace{0.3em} ``Grading the Grader: Lessons from Evaluating an Agentic Data Analysis System''}

\medskip
\textbf{Tian Zheng and Kai-Tai Hsu}\\
Department of Statistics\\
Columbia University\\
New York, NY
\end{center}

\medskip
\noindent This supplement provides additional details, tables, and case studies referenced in the main text.

%==============================================================
\section{Per-Step Instrumentation Details}
\label{supp:instrumentation}
%==============================================================

At every step, the wrapper records a structured \texttt{step\_data} object. Table~\ref{tab:step-fields} lists all recorded fields. Recording data at every step, rather than only the final state, allows us to detect cases where the correct answer was present in an early step but not surfaced as the final output.

\begin{table}[h]
\centering
\caption{Fields recorded in each \texttt{step\_data} object.}
\label{tab:step-fields}
\begin{tabular}{ll}
\toprule
\textbf{Field} & \textbf{Description} \\
\midrule
\texttt{instruction\_text}     & The instruction sent to the programmer \\
\texttt{instruction\_type}     & One of \texttt{original\_prompt}, \texttt{inspector\_revision}, \\
                               & \texttt{nudge\_original}, or \texttt{nudge\_generic} \\
\texttt{elapsed\_time}         & Wall-clock time for the step (seconds) \\
\texttt{token\_count}          & Token count via \texttt{tiktoken cl100k\_base} \\
\texttt{execution\_error}      & Boolean: did the code raise an error? \\
\texttt{scalar\_found}         & Boolean: was a numerical answer extracted? \\
\texttt{anchoring\_score}      & Keyword overlap with original question ($\in [0,1]$) \\
\texttt{execution\_scalar}     & Numerical value extracted from code output \\
\texttt{verbal\_scalar}        & Numerical value extracted from prose output \\
\texttt{answer\_region\_text}  & The output block selected as the answer region \\
\texttt{answer\_region\_score} & Keyword overlap score for the selected region \\
\texttt{candidate\_numbers}    & All numbers extracted from the answer region \\
\texttt{n\_candidates}         & Count of candidate numbers \\
\texttt{ambiguous\_answer}     & Boolean: more than 3 candidates (triggers nudge) \\
\bottomrule
\end{tabular}
\end{table}

%==============================================================
\section{Methods Reference}
\label{supp:methods-ref}
%==============================================================

Table~\ref{tab:methods-ref-supp} summarizes the key experimental settings used throughout the study.

\begin{table}[h]
\centering
\caption{Methods reference: experimental settings.}
\label{tab:methods-ref-supp}
\begin{tabular}{ll}
\toprule
\textbf{Setting} & \textbf{Value} \\
\midrule
Benchmark          & QRData (DSGym \citep{nie2026dsgym}) \\
Number of tasks $n$ & $153$ (all numerical QRData tasks) \\
Answer extraction   & Keyword-anchored region $+$ closest-match \\
Nudge modes        & A: answer format $+$ re-inject question; B: answer format only \\
Max retry turns    & $3$ \\
Max nudge & $2$\\
Per-task timeout & $300$ s \\
Outcome taxonomy   & clean / error\_recovered / nudge\_recovered / failed \\
Token counter      & \texttt{tiktoken cl100k\_base} (whitespace fallback) \\
Scalar tolerance   & $\mathrm{rtol}=0.02$, $\mathrm{atol}=10^{-6}$, $100\times$ for \% \\
Lenient grader      & GPT-4o-mini via OpenAI API, $3\%$ relative tolerance \\
\bottomrule
\end{tabular}
\end{table}

%==============================================================
\begingroup\color{black}
\section{Prompt Transparency and Grader Prompt Templates}
\label{supp:prompt-transparency}
%==============================================================

This section documents the wrapper-authored prompts used in Figure~1 (task packaging to LAMBDA, nudges, and the lenient grader). The goal is transparency and reproducibility: these prompts are not learned or optimized using the human labels, which are used only for post-hoc evaluation.

\noindent\textbf{Workflow alignment.} The prompt types below correspond to the Figure~1 steps: \emph{QRData task} $\rightarrow$ \emph{Nudge} $\rightarrow$ \emph{Lenient grader}.

\paragraph{Prompt layers.} LLM use occurs at multiple layers: (i) LAMBDA's internal programmer/inspector prompts; (ii) any DSGym-owned evaluation layers; and (iii) our wrapper-authored prompts that control nudging and lenient grading. We provide the wrapper-authored prompts verbatim below.

\paragraph{Nudge prompts.} Mode~A re-injects the original question together with an answer-format cue; Mode~B provides only the answer-format cue.

\noindent\textbf{Mode~A nudge template (re-inject question).}
\begin{verbatim}
Based on your analysis, please state a single numerical answer to the
following question.
Respond with ONLY the number, no explanation.

Question: {original_question}
\end{verbatim}

\noindent\textbf{Mode~B nudge template (format only).}
\begin{verbatim}
Based on your analysis, please state a single numerical answer to the
question.
Respond with ONLY the number, no explanation.
\end{verbatim}

\paragraph{Lenient grader prompts.} When strict grading does not succeed, the lenient grader first extracts a single numerical answer from the full transcript (or returns \texttt{UNCLEAR}) and then applies the $3\%$ tolerance rule.

\noindent\textbf{Lenient extraction template.}
\begin{verbatim}
You are grading the output of a data-analysis agent.
The agent was asked:

"{original_question}"

Here is the agent's full output:
---
{agent_output}
---

What single numerical value does the agent present as its final answer
to the question?
- If the agent explicitly states a final answer, extract that number.
- If the agent does not explicitly state a final answer but the output
  contains a computation result that answers the question, extract that
  number.
- If you cannot determine what the agent's answer is, respond with
  "UNCLEAR".

Respond with ONLY the number (or "UNCLEAR").
No explanation.
\end{verbatim}

\noindent\textbf{Lenient judge template (YES/NO fallback).}
\begin{verbatim}
Ground truth answer: {ground_truth}
Agent's extracted answer: {extracted_answer}

Are these the same value (within 3% relative tolerance)?
Respond YES or NO.
\end{verbatim}
\paragraph{Example (rendered nudge prompt).} For concreteness, we include an example of the rendered Mode~A nudge prompt for Case~2 (task \texttt{qrdata\_original\_qrdata\_35\_35}).

\begin{verbatim}
Based on your analysis, please state a single numerical answer to the
following question.
Respond with ONLY the number, no explanation.

Question: Use a Poisson distribution to approximate the data. What is
the event rate of the Poisson distribution? Please round to the
nearest tenth.
\end{verbatim}
A case-level ``LLM use report'' (including prompt IDs and the wrapper-authored
prompt texts used in the run) is included in the project's GitHub artifacts as:
\texttt{Results/case\_35\_35\_llm\_use\_report.md}.
\endgroup

%==============================================================
\section{Answer Extraction and Grading Details}
\label{supp:extraction-details}
%==============================================================

\paragraph{Heuristic parser (baseline).}
The baseline extraction heuristic---referred to as the ``heuristic parser'' throughout the paper---applies a last-number rule: it scans the agent's output from end to start and returns the first number matching a general numeric pattern (integers, decimals, scientific notation). This approach systematically produces false negatives in agentic outputs because the last number is often a step counter (e.g., ``1.~Visualize\ldots''), a confidence level (e.g., $\alpha = 0.05$), or an array index rather than the intended answer. 

\paragraph{Answer-region confidence score.}
During keyword-anchored extraction, the agent's output is segmented into blocks (split at blank lines or HTML boundaries). Each block is scored by keyword overlap with the original question plus a bonus for answer-indicator phrases (``the answer is,'' ``therefore,'' ``result''). The \emph{answer-region confidence score} is the highest block-level score; it ranges from $0$ (no keyword overlap) to values above $1$ (when indicator bonuses contribute). If no block scores above $0.1$, the pipeline falls back to the last block, flagged as low-confidence. In the full run, tasks in the highest-confidence band ($\ge 0.75$) achieve a $53\%$ strict-pass rate.

\paragraph{Strict grader: candidate selection.}
The strict grader receives all candidate numbers extracted from the answer region (after filtering out line numbers, array indices, and boolean-context values). It computes the absolute distance from each candidate to the ground truth, selects the closest, and checks whether the relative tolerance is below $\mathrm{rtol}=0.02$ (with $\mathrm{atol}=10^{-6}$; $100\times$ for \% answers):
\[
\text{rel\_tol} = \frac{|c_{\mathrm{closest}} - \text{GT}|}{\max(|\text{GT}|,\, 10^{-9})} < 0.02
\]
If multiple candidates are equidistant, the first encountered is selected. The selected candidate becomes the task's \texttt{strict\_matched\_value} regardless of whether it passes the tolerance check. If strict grading fails, the lenient grader is invoked on the full agent output (not the extracted candidates), making it architecturally independent of the parsing pipeline.

\begin{figure}[h]
\centering
\includegraphics[width=0.7\textwidth]{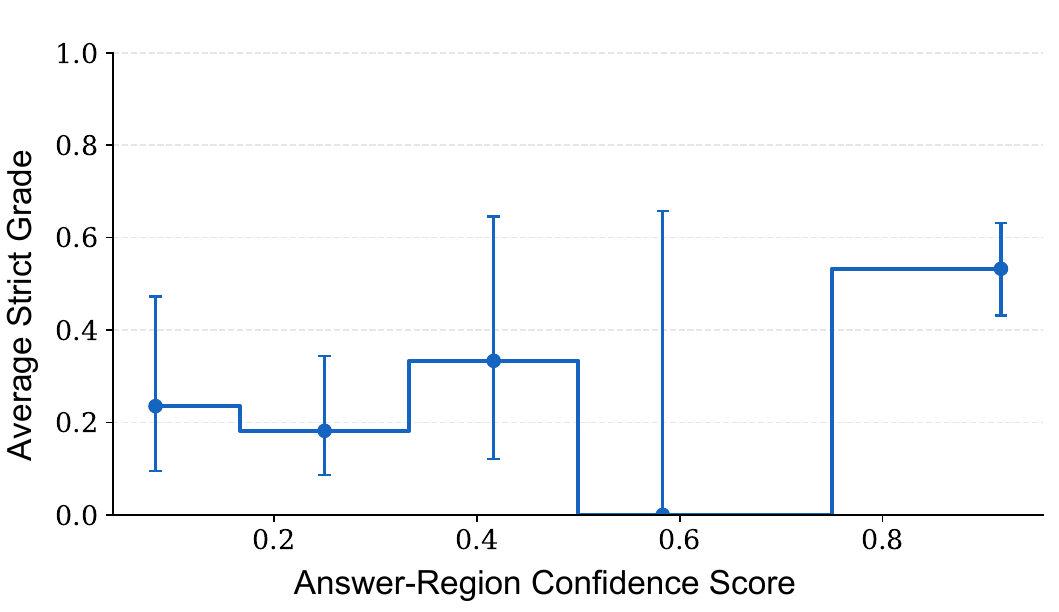}
\caption{Answer-region confidence score vs.\ average strict grade (Mode~A, $n=153$). Bars show binned average strict grades; error bars indicate $95\%$ binomial confidence intervals. Higher confidence regions are associated with higher strict-pass rates.}
\label{fig:region-confidence}
\end{figure}

\paragraph{Confidence score and average strict grade.}
Figure~\ref{fig:region-confidence} plots the average strict grade against the answer-region confidence score. Tasks in the highest-confidence band ($\ge 0.75$) achieve a $53\%$ average strict grade---roughly double the rate at lower confidence levels. The dip around $0.6$ reflects a small bin with few tasks. The overall trend confirms that the answer-region confidence score is a useful internal quality indicator: higher confidence signals that the extraction pipeline identified a keyword-rich answer region, which in turn increases the probability of a strict pass.

%==============================================================
\section{Case Study 1: Parsing Failure on a Clean Run}
\label{supp:case-clean}
%==============================================================

This case illustrates how a correct computation can be misgraded when the agent's output contains extraneous printed values that confuse the extraction pipeline.

\medskip
\noindent\textbf{Task ID:} \texttt{qrdata\_original\_qrdata\_17\_17} \quad \textbf{Variable type:} mixed \quad \textbf{Ground truth:} \texttt{0.61}

\smallskip
\noindent\textbf{Question:} \emph{``Compute the probability a randomly selected loan from the data set is for someone who has a mortgage or owns her home. Please round to the nearest hundredth.''}

\smallskip
\noindent\textbf{Step~0 (original prompt).} The programmer writes a single code block that (1)~identifies the \texttt{homeownership} column, (2)~filters on \texttt{MORTGAGE} and \texttt{OWN}, (3)~computes and prints the proportion, and (4)~calls \texttt{data.head()} to display the first five rows of the 55-column dataset. Anchoring $= 0.40$. The code runs without error.

\smallskip
\noindent\textbf{Execution output} (abridged):
\begin{verbatim}
Probability of a randomly selected loan being for someone
who has a mortgage or owns her home: 0.61

         emp_title  emp_length state homeownership  annual_income ...
0  global config..        3.0    NJ      MORTGAGE      90000.0  ...
1  warehouse off..       10.0    HI          RENT      40000.0  ...
  [5 rows x 55 columns]
\end{verbatim}

\noindent\textbf{Verbal text} (closing): \emph{``The probability \ldots{} is approximately 0.61 (61\%). \ldots{} Next, you can: Explore the relationship between homeownership status and loan default\ldots''}

\smallskip
\noindent\textbf{What went wrong.} The correct answer (0.61) appears clearly in both the execution output and the verbal text. However, the \texttt{data.head()} call floods the execution output with a 55-column table containing dozens of numeric values (employee lengths, income figures, debt ratios, balances). The answer-region selector scored the code block---which contains the \texttt{round(probability, 2)} expression---higher than the execution output block, and extracted the candidate number \texttt{2} (from the rounding argument). Mode~A prediction: \texttt{2.0}; strict match: \textbf{False}. The lenient GPT-4o-mini judge correctly identified \texttt{0.61} from the full output; lenient match: \textbf{True}.

\smallskip
\noindent\textbf{Mode~B comparison.} Under Mode~B, a different extraction path selected the execution output block, correctly extracting \texttt{0.61}; strict match: \textbf{True}. The same agent computation, graded by the same pipeline with a different run's output formatting, produces opposite strict verdicts.

\paragraph{Discussion.} This task exemplifies the ``hard to grade, easy to execute'' regime. The agent's analytical reasoning is trivially correct---a single filter-and-mean computation---but the extraneous \texttt{data.head()} output creates an extraction challenge. The divergence between Mode~A (strict-wrong) and Mode~B (strict-pass) on the \emph{same question with the same ground truth} illustrates how fragile automated grading is: a cosmetic difference in output formatting flips the verdict. The lenient grader catches the error in Mode~A, demonstrating the value of the multi-tier grading cascade, but also underscoring that the strict score alone is an unreliable measure of agent capability.

%==============================================================
\section{Case Study 2: Design Mismatch and Nudge Recovery}
\label{supp:case-nudge}
%==============================================================

This case illustrates the most common nudge scenario: the agent computes the correct answer on the first pass but does not present it as a final scalar, instead appending exploratory suggestions that prevent extraction.

\medskip
\noindent\textbf{Task ID:} \texttt{qrdata\_original\_qrdata\_35\_35} \quad \textbf{Variable type:} continuous \quad \textbf{Ground truth:} \texttt{4.4}

\smallskip
\noindent\textbf{Question:} \emph{``Use a Poisson distribution to approximate the data. What is the event rate of the Poisson distribution? Please round to the nearest tenth.''}

\smallskip
\noindent\textbf{Step~0 (original prompt).} The programmer loads \texttt{ami\_occurrences.csv} and displays the first few rows, requesting clarification on which column to model. Anchoring $= 0.435$. No answer is extracted (\texttt{scalar\_found = False}); the execution output contains only the raw data table. The verbal text closes with: \emph{``Next, you can: Calculate and round the event rate (mean of the `ami' column)\ldots''}

\smallskip
\noindent\textbf{Step~1 (first nudge).} Mode~A re-injects the original question with a answer format. The programmer computes \texttt{data['ami'].mean()} and obtains $\lambda = 4.4$, printed as \texttt{np.float64(4.4)}. However, the verbal text again appends three numbered suggestions (``1.~Visualize \ldots{} 2.~Perform a goodness-of-fit test \ldots{} 3.~Explore other distributions\ldots''). The answer region now contains 9 candidate numbers (including the step numbers 1, 4, and repeated instances of 4.4), triggering the ambiguity condition. \texttt{scalar\_found = False}. Under Mode~B (format-only nudge), anchoring drops to $0.0$, but the same ambiguity occurs.

\smallskip
\noindent\textbf{Step~2 (second nudge).} Both modes re-issue the nudge. The programmer responds with just \texttt{4.4}---a single number with no surrounding text. \texttt{scalar\_found = True}; strict match: \textbf{True}.

\paragraph{Discussion.} The correct answer was present in Step~1's execution output. The two nudges were not correcting an analytical error; they were providing the answer template cue that LAMBDA's conversational interface does not naturally produce. LAMBDA's programmer is explicitly prompted to end each turn with ``suggestions for next steps'' \citep[Figure~5]{sun2026lambda}---natural in conversation, but adversarial for single-shot benchmark extraction. This design-mismatch pattern accounts for the majority of nudge-recovered tasks in the full run (94 of 98 nudged tasks recovered under Mode~A).

The Mode~A/B comparison is instructive: both modes reach the correct answer in the same number of steps, but anchoring diverges---Mode~A maintains $0.435$ throughout (the question is re-injected each time), while Mode~B drops to $0.0$ after the first nudge (the agent receives only a format cue with no reminder of the original question). On this task the anchoring difference is inconsequential, but on harder tasks where the agent's reasoning has genuinely drifted, the re-injection in Mode~A can be the difference between recovery and failure.

%==============================================================
\section{Case Study 3: Statistical Error Invisible to Automated Grading}
\label{supp:case-failed}
%==============================================================

This case illustrates a genuine analytical error that no layer of automated grading detects---a failure mode that is easy to grade but hard to execute correctly.

\medskip
\noindent\textbf{Task ID:} \texttt{qrdata\_original\_qrdata\_49\_49} \quad \textbf{Variable type:} categorical \quad \textbf{Ground truth:} \texttt{0.012}

\smallskip
\noindent\textbf{Question:} \emph{``Given the hypothesis that the higher-quality blades will pass inspection 3\% more frequently than the standard-quality blades, calculate the p-value for the hypothesis. Please round to 3 decimal places.''}

\smallskip
\noindent\textbf{Step~0 (original prompt).} The programmer loads \texttt{drone\_blades.csv} and displays the first rows, identifying two supplier groups and a pass/fail inspection column. Anchoring $= 0.378$. No answer is extracted; the output contains only the data preview.

\smallskip
\noindent\textbf{Step~1 (first nudge, Mode~A).} The programmer writes a hypothesis test. The code computes the observed difference in pass rates, constructs a $z$-statistic under the null $H_0: p_2 - p_1 = 0.03$, and calculates:
\begin{verbatim}
z_stat = (p_diff - 0.03) / se
p_value = 1 - norm.cdf(z_stat)
\end{verbatim}
This is a one-sided upper-tail test. The execution output is \texttt{np.float64(0.006)}, but the code raises an error on a subsequent line. The answer region contains 41 candidate numbers (proportions, counts, the hypothesis value 0.03, and the p-value 0.006), triggering the ambiguity condition. Anchoring remains at $0.378$ (Mode~A re-injected the question).

\smallskip
\noindent\textbf{Steps~2--3 (inspector revisions).} The inspector detects the execution error and issues revision suggestions. The programmer rewrites the code, but makes the same statistical choice: \texttt{p\_value = 1 - norm.cdf(z\_stat)} $= 0.006$. The code now runs without error. The answer region narrows to 2 candidates ($0.006$ and $3.0$); the extraction pipeline selects $0.006$ as the closest match to the question's context. Final prediction: \texttt{0.006}; strict match: \textbf{False}; lenient match: \textbf{False}.

\smallskip
\noindent\textbf{The error.} The ground truth is $0.012$, exactly twice the agent's answer. The question asks for ``the p-value for the hypothesis,'' which is conventionally a two-sided test: $p = 2 \times P(Z > |z|)$. The agent computed the one-sided upper-tail probability. This is a substantive statistical reasoning difference---choosing a different test specification---not a coding bug or parsing failure. The code runs cleanly, the inspector confirms execution without error (statistical logic is outside its scope), and the verbal text confidently states: \emph{``The p-value of 0.006 strongly suggests that the higher-quality blades pass inspection at least 3\% more frequently.''}

\smallskip
\noindent\textbf{Mode~B comparison.} Under Mode~B, the trajectory is similar: the agent produces $0.006$ after two nudges. Anchoring drops to $0.0$ (no question re-injection), but the outcome is the same, confirming there is no drifting.

\paragraph{Discussion.} This task is the mirror image of Case~1. Where the loan-probability task was easy to execute but hard to grade, this task is easy to grade---the agent produces a single clean number---but hard to execute precisely to match the ground truth. The evaluation framework functions as designed: the extraction pipeline finds the answer, the strict grader correctly rejects it, and the lenient grader also correctly rejects it. But no automated layer is available to unveil the workflow choice difference. The one-sided vs.\ two-sided distinction requires understanding the statistical semantics of the question, which lies beyond the scope of any scalar comparison.

This failure mode is particularly insidious because the agent's answer ($0.006$) is numerically close to the ground truth ($0.012$) and is statistically defensible. Only human inspection, guided by domain knowledge, can identify the root cause. The case illustrates why scalar benchmarks, however well-instrumented, are insufficient for evaluating statistical reasoning: they can detect \emph{that} an answer is not matched but not \emph{why}, and the ``why'' is essential for improving the system.

%==============================================================
\section{Mode~A vs.\ Mode~B Detailed Comparison}
\label{supp:mode-comparison}
%==============================================================

Table~\ref{tab:mode-comparison-supp} compares the two nudge modes across grading pipeline dynamics and grading outcomes. Mode~A re-injects the original question alongside the answer format; Mode~B uses the answer format alone.

\begin{table}[h]
\centering
\caption{Mode~A vs.\ Mode~B comparison ($n=153$).}
\label{tab:mode-comparison-supp}
\begin{tabular}{lcc}
\toprule
\textbf{Metric} & \textbf{Mode A} & \textbf{Mode B} \\
\midrule
Strict-pass rate         & 40.5\% & 43.1\% \\
Lenient-pass rate        & 45.8\% & 46.4\% \\
Nudge triggered             & 98   & 106   \\
Nudge recovered             & 94   & 98   \\
\texttt{clean}              & 31   & 26   \\
\texttt{error\_recovered}   & 24   & 21   \\
%\texttt{nudge\_recovered}   & 94   & 98   \\
%\texttt{failed}             & 4   & 8   \\
\bottomrule
\end{tabular}
\end{table}

\begin{table}[t]
\centering
\caption{Mode A vs Mode B comparison.}
\label{tab:mode-comparison}
\begin{tabular}{lccc}
\toprule
Metric & Mode A & Mode B & Mode A and B \\
\midrule
Nudges triggered (total) & 170 & 175 &  \\
%Scalar found & 153/153 (100.00\%) & 153/153 (100.00\%) &  \\
%Scalar found and agrees with GT & 62/153 (40.52\%) & 67/153 (43.79\%) & 118/153 (77.12\%) \\
Agrees with GT; Strict Grader & 62/153 (40.52\%) & 66/153 (43.14\%) & 47/153 (30.07\%) \\
Agrees with GT; Lenient Grader & 70/153 (45.75\%) & 71/153 (46.41\%) & 55/153 (35.95\%)\\
%Nudge-count agreement:& \multicolumn{3}{l}{ 92/153 (60.13\%)} \\
%Scalar-found agreement:& \multicolumn{3}{l}{ 153/153 (100.00\%)} \\
%Between-mode agreement; Strict grader& \multicolumn{3}{l}{ 119/153 (77.78\%)} \\
%Between-mode agreement; Lenient grader& \multicolumn{3}{l}{ 122/153 (79.74\%)} \\
%Scalar-value agreement (both found): & \multicolumn{3}{l}{83/153 (54.25\%)} \\
\bottomrule
\end{tabular}
\end{table}

%==============================================================
\section{Verbal--Execution Disagreement Examples (Heuristic Parser)}
\label{supp:disagreements}
%==============================================================

Table~\ref{tab:disagreements-supp} presents selected tasks where the verbal and execution scalars disagree under the \emph{heuristic parser}. The keyword-anchored parser eliminates all verbal--execution disagreements (see main text); the examples here illustrate why the heuristic approach fails and motivate the switch to keyword-anchored extraction. In all rows, the execution scalar matches the ground truth; the verbal scalar is wrong due to extraction artifacts of the heuristic parser.

\begin{table}[h]
\centering
\small
\caption{Selected verbal--execution disagreements under the heuristic parser. Execution scalar matches ground truth in all rows.}
\label{tab:disagreements-supp}
\begin{tabular}{p{4.2cm}rrrl}
\toprule
\textbf{Question (trunc.)} & \textbf{GT} & \textbf{Exec} & \textbf{Verbal} & \textbf{Parsing error} \\
\midrule
Proportion w/ stroke\dots     & $0.20$  & $0.20$  & $2.00$  & step number   \\
Median AQI value\dots         & $30$    & $30.00$ & $91.00$ & wrong number  \\
$95\%$ CI for $p$\dots        & $0.844$ & $0.843$ & $0.05$  & conf.\ level  \\
$p$-value for hypothesis\dots & $0.865$ & $0.870$ & $4.00$  & step number   \\
\bottomrule
\end{tabular}
\end{table}

\end{document}